\documentclass[10pt, a4paper]{article}
\usepackage{lrec-coling2024}
\usepackage{graphicx}
\usepackage{amsmath}
\usepackage{colortbl}
\usepackage{pifont}
\usepackage{multirow}

\usepackage[utf8]{inputenc}
\usepackage{microtype}
\usepackage{inconsolata}
\usepackage{colortbl}
\usepackage{url}

\title{Navigating Prompt Complexity for Zero-Shot Classification: A Study of Large Language Models in Computational Social Science}

\name{\begin{tabular}{c}Yida Mu, Ben P. Wu, William Thorne, Ambrose Robinson, \\ Nikolaos Aletras, Carolina Scarton, Kalina Bontcheva, Xingyi Song \end{tabular} \\}

\address{Department of Computer Science, The University of Sheffield \\
         \{y.mu, bpwu1, wthorne1, arobinson10, n.aletras, c.scarton, k.bontcheva, x.song\}@sheffield.ac.uk\\}

\abstract{
Instruction-tuned Large Language Models (LLMs) have exhibited impressive language understanding and the capacity to generate responses that follow specific prompts. However, due to the computational demands associated with training these models, their applications often adopt a zero-shot setting. In this paper, we evaluate the zero-shot performance of two publicly accessible LLMs, ChatGPT and OpenAssistant, in the context of six Computational Social Science classification tasks, while also investigating the effects of various prompting strategies. Our experiments investigate the impact of prompt complexity, including the effect of incorporating label definitions into the prompt; use of synonyms for label names; and the influence of integrating past memories during foundation model training. The findings indicate that in a zero-shot setting, current LLMs are unable to match the performance of smaller, fine-tuned baseline transformer models (such as BERT-large). Additionally, we find that different prompting strategies can significantly affect classification accuracy, with variations in accuracy and F1 scores exceeding 10\%.
 \\ \newline  }

\begin{document}

\maketitleabstract

\section{Introduction}
Instruction fine-tuning  \citep{ouyangTrainingLanguageModels2022} has facilitated transfer learning for Large Language Models (LLMs) to unseen tasks at scale. To leverage LLMs as versatile natural language processors, there is an immediate effort to ascertain their zero-shot performance on challenging tasks. Social media analysis is an active area of research with a number of complex, domain-specific tasks which can be utilised for harm reduction \citep{waseem2017understanding} and preventing the spread of misinformation \citep{zubiaga2018detection}. LLMs have great potential to assist with such computational social science (CSS) tasks, both in automatic data annotation and social media analysis \citep{kuzman2023chatgpt,reiss2023testing,törnberg2023chatgpt4}. 
Hence, it is important to understand the capabilities and limitations of the latest instruction fine-tuned LLMs for addressing such CSS tasks. In this paper, we are primarily focusing on answering the following research questions (RQ):\footnote{Accepted at LREC-COLING 2024.} 

\begin{itemize}
    \item \textbf{(RQ 1)} What level of zero-shot performance can LLMs achieve in social media classification tasks? How does zero-shot LLM  performance compare against smaller state-of-the-art language models fine-tuned to the specific analysis task?

    \item \textbf{(RQ 2)} What are the most effective LLM prompt strategies for social media classification tasks in a zero-shot setting?

    \item \textbf{(RQ 3)} \textcolor{black}{Was the pre-training corpus of the large model already inclusive of these datasets prior to the experiment (i.e., data leakage issues)?}
\end{itemize}

To answer those research questions, we conduct a series of controlled experiments to investigate the zero-shot performance of two off-the-shelf instruction fine-tuned large language models using different prompting strategies. Namely, we experiment with GPT-3.5-turbo (GPT),\footnote{\url{https://openai.com/blog/chatgpt}} the most widely used proprietary instruction fine-tuned large language model; and OpenAssistant-LLaMA (LLaMA-OA) \cite{kopf2023openassistant}, an open source LLM instruction fine-tuned based LLaMA \cite{touvron2023llama}. We use six social media analysis NLP tasks to evaluate the classification performance of LLMs using different prompt complexity levels (including providing few-shots examples and publication information of benchmark datasets in the prompt). The findings are also compared against baselines employing standard techniques such as fine-tuning BERT.

It must be noted that the scope of this paper is on evaluating the performance of off-the-shelf, instruction fine-tuned language models on social media classification tasks, in a zero-shot setting. The evaluation of  foundation language models without instruction fine-tuning is out of the scope of this paper.

Our main findings are:
\begin{itemize}
\item \textbf{(i)} Task-specific fine-tuned models still generally tend to outperform LLMs in most zero-shot settings, even when the fully fine-tuned model (e.g., BERT-large model) is significantly smaller.

\item \textbf{(ii)} Using prompting ensemble methods (e.g., on synonyms) can increase the performance and robustness of LLMs.

\item \textbf{(iii)} Detailed and complex prompting strategies are not necessary.
\end{itemize}

\section{Related Work}
Both models evaluated in this work, GPT (also referred to as ChatGPT) and LLaMA-OA, have been trained using Reinforcement Learning with Human Feedback (RLHF) in conjunction with instruction tuning, as first explored in \citet{ouyangTrainingLanguageModels2022}. Instruction tuning is the fine-tuning of language models on NLP tasks rephrased as instructions and prior work has shown that it is an effective way of training LLMs to perform zero-shot on unseen tasks. \cite{weiFINETUNEDLANGUAGEMODELS2022, sanhMultitaskPromptedTraining2022} \citet{longpreFlanCollectionDesigning2023} carried out a detailed ablation study on non-RLHF instruction tuning methods across the general NLP tasks in the Flan 2022 collection and found that T5 instruction tuned on the Flan performed surprisingly well on held-out tasks when compared to models directly fine-tuned on said task. Tuning with human feedback could be the next step in improving instruction tuning in this area.

\citet{ziems2023can} sets a roadmap for employing LLMs as data annotators by establishing prompting best practices and an evaluation of the zero-shot performance of 13 language models on 24 tasks in computational social sciences. \textcolor{black}{In the financial domain, \citep{li2023chatgpt} reveal that ChatGPT and GPT-4 outperform the performance of supervised models, which have been fine-tuned with domain-specific data, in several financial benchmarks.}

To evaluate the zero-shot performance of ChatGPT for text classification, \citet{kuzman2023chatgpt} compares against a fine-tuned XLM-RoBERTa model for the task of automatic genre classification in English and Slovenian. They show that ChatGPT outperforms the baseline on unseen datasets and that there is no drop in performance when provided with Slovenian examples. \citet{ganesan2023systematic} use Facebook posts to classify user personality traits, based on openness, conscientiousness, extroversion, agreeableness, and neuroticism. They find that GPT-3 performs poorly on binary and worse yet on tertiary ranking for each trait.

LLMs have also been applied in mental health applications. \citet{lamichhane2023evaluation} evaluate ChatGPT's ability to classify stress, depression, and suicidal inclination from Reddit posts. Although ChatGPT significantly outperforms their baseline, the baseline consisted of a simple prediction of the majority class.

For toxicity detection, \citet{wang2022toxicity} analyse GPT-3's generative and discriminative zero-shot capabilities, finding that performance is only slightly better than a random baseline. However, the authors argue that the generative task allows for nuanced distinction of toxicity in the, somewhat subjective, binary setting.

\citet{törnberg2023chatgpt4} find that ChatGPT-4 outperforms non-expert annotators in identifying the political affiliation of Democratic or Republican party members based on their tweets during the 2020 US election. \citet{wu2023large} use ChatGPT to rank the conservatism of representatives in the 116th US Congress through a series of pairwise match ups, showing a high correlation with DW-NOMINATE scores.

As LLMs improve their performance on language generation tasks, the risk of misinformation and propaganda increases. \citet{mitchell2023detectgpt} propose DetectGPT, a perturbation-based zero-shot method for identifying machine-generated passages. \cite{su2023detectllm} further develop this approach with DetectLLM-LRR and -NPR, achieving improved efficiency and improved performance respectively.

\textcolor{black}{Note that our work is distinct from previous research \citep{ziems2023can}; we evaluate Large Language Models (LLMs) on a different set of benchmarks and experiment with various prompt modification strategies, including replacing original labels with synonyms and incorporating arXiv paper titles.}

\section{Methodology}

\begin{table*}[!t]
\resizebox{\textwidth}{!}{%
\begin{tabular}{|l|l|}
\hline
\rowcolor[HTML]{DAE8FC} \textbf{Task} &
  \textbf{Basic} \\ \hline
\textit{Bragging} &
  \begin{tabular}[c]{@{}l@{}} \textbf{Basic Instruction} (i.e., Identify whether or not a tweet includes a bragging statement.) \\ \textbf{+ Constraints} (i.e., Only reply (bragging) or (not bragging).)\\ \textbf{+ Text} (e.g., Tweet: Come watch me and @USER face off in 2K best of 3 series \#braggingrights @USER\\ you next boiiii.) \end{tabular} \\ \hline
\rowcolor[HTML]{DAE8FC} \textbf{Task} &
  \textbf{Basic + T/L Desc} \\ \hline
\textit{Vaccine} &
  \begin{tabular}[c]{@{}l@{}} \textbf{Basic Instruction}\\ +\textbf{ T/L Desc}  Tweets that have been assigned to the class `pro vaccine' express a positive opinion regarding the \\ vaccination. Tweets belonging to the `anti vaccine' class express a negative opinion towards COVID-19 \\ vaccination. The `neutral' class mainly includes news related to the development of vaccines, tweets that do \\not express  a clear opinion, such as questions regarding the vaccine, informative tweets concerning vaccination. \\\bf + Constraints + Text\end{tabular} \\ \hline
\rowcolor[HTML]{DAE8FC} \textbf{Task} &
  \textbf{Few-sample} \\ \hline
Complaint &
  \begin{tabular}[c]{@{}l@{}} \textbf{Basic Instruction} \\ \textbf{+ Few-samples} (e.g., (i) Complaint: @USER @USER give the timeline by which I'll receive my cashback\\ which I should have received by 15th October 2017. (ii) Not Complaint: I just gave 5 stars to Nancy at @USER \\for the great service I received!) \\\bf + Constraints + Text\end{tabular} \\ \hline
\rowcolor[HTML]{DAE8FC} \textbf{Task} &
  \textbf{Memory Recall} \\ \hline
Hate Speech &
  \begin{tabular}[c]{@{}l@{}} \textbf{Basic Instruction}\\ \textbf{ + arXiv Paper Title} (i.e., Recall this paper: Hateful symbols or hateful people? predictive features for\\ hate speech detection on twitter.)
  \\\bf + Constraints + Text\end{tabular} \\ \hline
\end{tabular}
}
\caption{Prompt examples across different settings.}
\label{tab:promptssss}
\end{table*}

\subsection{Prompting Strategies}

Following the prompting approaches described by \citet{child2019generating,ziems2023can}, we develop prompts by (i) adding instructions after the context (e.g., task description) and (ii) using constraints (e.g., `Only reply with Bragging' or `Not Bragging.') at the end. We observe that using constraints can effectively avoid cases of model uncertainty (e.g., `As an AI model, I cannot answer this question.') and guide models to generate the expected outputs.

For consistency, we use the same prompts for both GPT and LLaMA-OA. Examples of different prompt strategies are displayed in Table \ref{tab:promptssss}. To examine the zero-shot predictive performance of LLMs, we carry out a comprehensive set of experiments using four different prompting strategies.

\paragraph{\textbf{Basic Instruction (Basic)}:} We only provide a basic instruction without including detailed task and label descriptions. For example, for the bragging detection task, our prompt is: \textit{`Identify whether or not a tweet includes a bragging statement. + Constraints + Text'.} Two possible configurations are tested, namely adding the prompt \textit{before} or \emph{after} the text respectively.

\paragraph{\textbf{Task and Label Description (T/L Desc)}:} Building upon the Basic Instruction Round, we provide additional information in the prompt by including task and label descriptions (see Table \ref{tab:promptssss}). Note that we use the labels and task descriptions detailed in the original papers on the respective datasets. The format of the prompts used for the Task and Label Description Round is: \textit{`Basic Instruction + Task and Label Descriptions + Constraints + Text'}.

\paragraph{Few-sample Prompting (Few-sample):}
We also test a few-sample prompting strategy by adding one example selected from the training set for each label. The prompt designed for the few-sample experiments is: \textit{`Basic Instruction + Few-shot Examples + Constraints + Text'}. Note that using few-sample as input is still a type of zero-shot setup, as we do not fine-tune the model.

\paragraph{\textbf{Memory Recall (Recall)}:}
We observe that both GPT and LLaMA-OA can recall papers published before September 2021. Since arXiv papers are part of the training data of the LLMs, we also include the title of the source paper in the prompt when evaluating the model's zero-shot performance. For example, we include paper information by using this prompt: \textit{`Recall this paper [Paper Title] + Basic Instruction + Constraints + Text'}. For such recall prompts, we only perform experiments on datasets published before September 2021. For reference, we examine the variations in performance across different checkpoints to assess whether instruction fine-tuning might influence the efficacy of the classification task.

\subsection{Synonyms} 
\textcolor{black}{LLMs might generate different outputs when using prompts which are semantically similar} (e.g., synonyms\footnote{Appropriate synonyms were selected by consulting \url{https://www.thesaurus.com}.}). 
To test the generalisability of LLMs, we substitute the names of each class with words that have the same or similar meaning. For example, we test the synonyms `hateful', `toxic', and `abusive' to replace the original category `offensive'. We also use two ensemble learning approaches to improve predictive performance by combining the outputs from all synonyms settings for each dataset:
\begin{itemize}
    \item \textbf{Ensemble Majority:} 
    We select the category that has been selected the most times across all synonym experiments.  
    \item \textbf{Ensemble All Agreed:} 
    We also experiment with a stricter setting that considers only model outputs that are in the same category (i.e., Complaint, Criticism, dissatisfaction, etc.) using all synonyms.
    For example, we consider the LLM that uses all synonyms predicted as complaints, otherwise they are considered non-complaints. We only report this metric for datasets with binary classes.
\end{itemize}

\section{Data}
\begin{table*}[!t]
\centering
\resizebox{\textwidth}{!}{%
\begin{tabular}{|l|c|c|}
\hline
\rowcolor[HTML]{DAE8FC} 
\textbf{Dataset} & \multicolumn{1}{l|}{\textbf{\# of Posts}} & \textbf{Class (\# of Posts)}  \\ \hline
\textit{Rumour Stance}  & 5,568  & Support (1,004) / Deny (415) / Query (464) / Comment (3,685)   \\ \hline
\textit{Vaccine Stance} & 2,792  & Pro Vaccine (991) / Anti Vaccine (791) / Neutral (1,010)       \\ \hline
\textit{Complaint}  & 3,449  & Complaint (1,232) / Not Complaint (2,217)                      \\ \hline
\textit{Bragging}   & 6,696  & Bragging (781) / Not Bragging (5,915)                          \\ \hline
\textit{Sarcasm}   & 4,868  & Sarcasm (1,067) / Not Sarcasm (3,801)                         \\ \hline
\textit{Hate speech} & 16,907 & Offensive (5,348)  / Non-offensive (11,559)                                      \\ \hline
\end{tabular}
}
\caption{Dataset Specifications.}
\label{tab:benchmarks}
\end{table*}
In order to ensure a comprehensive evaluation of LLM performance, we select six datasets that cover a wide range of computational social science tasks and different time spans. In particular, some of them were created before September 2021, while others were collected after the release of the LLMs used in this paper.
%This paper evaluates the zero-shot classification performance of LLMs on six NLP tasks in computational social science. 
All datasets are in English with manually annotated class labels. We detail dataset specifications and statistics in Table \ref{tab:benchmarks}:

\begin{itemize}
    \item \textbf{Complaint} 
    This task aims to identify whether a tweet expresses a complaint, which is defined as `a negative mismatch between reality and expectations in a particular situation' (e.g., customer complaints on Twitter) \citep{olshtain198710}.
    We use a dataset developed by \citet{preoctiuc2019automatically}  consisting of 3,449 English tweets annotated with one of two categories, i.e., complaints or not complaints.
    
    \item \textbf{Vaccine Stance} This task aims to automatically predict the stance of tweets towards COVID-19 vaccination \citep{cotfas2021longest,mu2023vaxxhesitancy}.
    The dataset developed by \citep{cotfas2021longest} provides 2,792 tweets belonging to one of three stance categories: pro-vaccine, anti-vaccine, or neutral. 

    \item \textbf{Bragging} 
    This task aims to classify whether a tweet is bragging or not bragging. We evaluate on a dataset developed by \citet{jin2022automatic} which contains 6,696 tweets labelled as either bragging or not bragging.

    \item \textbf{Rumour Stance} 
    We use the RumorEval 2017 dataset which is developed by \citet{derczynski-etal-2017-semeval}.
    Here, we use the dataset for 4-way rumour stance classification, i.e., determining the stance of a reply towards a given source post (i.e. rumour) as either supporting, denying, questioning, or commenting.

    \item \textbf{Sarcasm} 
    The sarcasm detection task is to identify whether a given tweet is intended to be sarcastic or not. We evaluate the task on the Semeval-2022 Task 6 dataset \citep{abu-farha-etal-2022-semeval}, which contains 4,868 tweets labelled as either sarcasm or non-sarcasm.

    \item \textbf{Hate Speech} 
    The task of hate speech detection aims to study anti-social behaviours, e.g., racism and sexism in social media.
    We evaluate on a dataset developed by \citet{waseem-hovy:2016:N16-2} with a binary classification setup, i.e., offensive or non-offensive.
\end{itemize}

\section{Experimental Setup}
\subsection{Large Language Models}
Our experiments are conducted using two publicly accessible large language models: %GPT-3.5-turbo (GPT) and OpenAssistant-LLaMa (OA).

\paragraph{GPT-3.5-turbo (GPT)}\footnote{\url{https://platform.openai.com/docs/models/gpt-3-5}} is an enhanced version of the GPT-3 language model with instruction fine-tuning.  
GPT can be employed for a wide range of NLP tasks, including machine translation, common sense reasoning, and question answering. The experiments use the GPT model via the official OpenAI API.\footnote{\url{https://platform.openai.com/docs/api-reference}}

\paragraph{LLaMA-OA}
We employ the LLaMA-OA model developed by LAIONAI,\footnote{\url{https://laion.ai/}} which fine-tunes the vanilla LLaMA \citep{touvron2023llama} 30B model using the OpenAssistant dataset \citep{kopf2023openassistant}. Since the original LLaMA models are not allowed to be shared by individuals, LAIONAI could not release the weights for LLaMA-OA on huggingface but released xor (i.e., `Exclusive Or') weights\footnote{We use the \textsc{oasst-sft-6-llama-30b} version of the model. The xor weights can be found at: \url{https://huggingface.co/OpenAssistant/oasst-sft-7-llama-30b-xor}} applied to the original LLaMA weights and the check sum calculations performed to validate the conversion. In order to be able to run the experiments locally under hardware constraints, we applied 8-bit quantisation at model load time via BitsAndBytes \cite{dettmers20228bit} to decrease the inference memory requirements. 

\subsection{Baselines}
The zero-shot classification performance of the two LLMs is compared against a weak Logistic Regression baseline and a strong fully fine-tuned BERT-large baseline:

\paragraph{Logistic Regression}
We represent the text using TF-IDF and consider tokens that appear more than 5 times.

\paragraph{BERT-large}
We fine-tune BERT-large\footnote{\url{https://huggingface.co/bert-large-uncased}} \citep{devlin2019bert} by adding a linear classifier on top of the 24-layer transformer blocks. The special token `[CLS]' is used as the representation of each text.

\subsection{Data Splits}
For each benchmark task, we divide the dataset into training (80\%) and test (20\%) sets using stratified random splits\footnote{To generate class-stratified subsets, we employ a dataset split tool from \url{https://scikit-learn.org/stable/modules/generated/sklearn.model_selection}}.
The training set is used for supervised fine-tuning, and is further sub-divided into a training and a validation subsets (in a 3:1 ratio) for hyperparameter tuning (e.g., early stopping) purposes. Subsequently, we evaluate the performance of the fine-tuned baselines and zero-shot LLMs on the 20\% test set.

\begin{table*}[!t]
\small
\centering
\resizebox{\textwidth}{!}{%
\begin{tabular}{|l|cc|cc|cc|}
\hline
\rowcolor[HTML]{DAE8FC} 
\multicolumn{1}{|c|}{\cellcolor[HTML]{DAE8FC}} &
  \multicolumn{2}{c|}{\cellcolor[HTML]{DAE8FC}\textbf{Complaint}} &
  \multicolumn{2}{c|}{\cellcolor[HTML]{DAE8FC}\textbf{Vaccine Stance}} &
  \multicolumn{2}{c|}{\cellcolor[HTML]{DAE8FC}\textbf{Bragging}} \\ \cline{2-7} 
\rowcolor[HTML]{DAE8FC} 
\multicolumn{1}{|c|}{\multirow{-2}{*}{\cellcolor[HTML]{DAE8FC}\textbf{Model}}} &
  \multicolumn{1}{c|}{\cellcolor[HTML]{DAE8FC}\textbf{Accuracy}} &
  \cellcolor[HTML]{DAE8FC}\textbf{F1} &
  \multicolumn{1}{c|}{\cellcolor[HTML]{DAE8FC}\textbf{Accuracy}} &
  \textbf{F1} &
  \multicolumn{1}{c|}{\cellcolor[HTML]{DAE8FC}\textbf{Accuracy}} &
  \textbf{F1} \\ \hline
\rowcolor[HTML]{DAE8FC} 
\textit{Logistic Regression} &
  \multicolumn{1}{c|}{\cellcolor[HTML]{DAE8FC}81.4} &
  79.7 &
  \multicolumn{1}{c|}{\cellcolor[HTML]{DAE8FC}72.8} &
  73.1 &
  \multicolumn{1}{c|}{\cellcolor[HTML]{DAE8FC}88.6} &
  58.8 \\ \hline
\rowcolor[HTML]{DAE8FC} 
\textit{BERT-large} &
  \multicolumn{1}{c|}{\cellcolor[HTML]{DAE8FC}89.4} &
  88.6 &
  \multicolumn{1}{c|}{\cellcolor[HTML]{DAE8FC}\textbf{81.5}} &
  \textbf{81.3} &
  \multicolumn{1}{c|}{\cellcolor[HTML]{DAE8FC}\textbf{91.3}} &
  \textbf{76.1} \\ \hline
\textit{GPT Basic After} &
  \multicolumn{1}{c|}{84.9} &
  84.1 &
  \multicolumn{1}{c|}{65.5} &
  65.8 &
  \multicolumn{1}{c|}{81.1} &
  62.7 \\ \hline
\textit{GPT Basic Before} &
  \multicolumn{1}{c|}{\cellcolor[HTML]{C2F3D3}\textbf{89.7}} &
  \cellcolor[HTML]{C2F3D3}\textbf{88.7} &
  \multicolumn{1}{c|}{72.4} &
  73.6 &
  \multicolumn{1}{c|}{84.3} &
  66.2 \\ \hline  
\textit{GPT T/L Desc} &
  \multicolumn{1}{c|}{89.0} &
  88.0 &
  \multicolumn{1}{c|}{73.3} &
  73.7 &
  \multicolumn{1}{c|}{84.9} &
  67.4 \\ \hline
\textit{GPT Memory Recall } &
  \multicolumn{1}{c|}{87.1} &
  86.4 &
  \multicolumn{1}{c|}{66.2} &
  66.9 &
  \multicolumn{1}{c|}{79.8} &
  64.6 \\ \hline
\textit{GPT Few-sample} &
  \multicolumn{1}{c|}{85.6} &
  85.2 &
  \multicolumn{1}{c|}{68.2} &
  69.4 &
  \multicolumn{1}{c|}{77.3} &
  61.8 \\ \hline  
\textit{LLaMA-OA Basic After} &
  \multicolumn{1}{c|}{65.5} &
  65.4 &
  \multicolumn{1}{c|}{60.5} &
  57.8 &
  \multicolumn{1}{c|}{57.8} &
  50.1 \\ \hline  
\textit{LLaMA-OA Basic Before} &
  \multicolumn{1}{c|}{80.1} &
  79.9 &
  \multicolumn{1}{c|}{64.2} &
  63.7 &
  \multicolumn{1}{c|}{82.8} &
  62.6 \\ \hline
\textit{LLaMA-OA Basic (OAT 7)} &
  \multicolumn{1}{c|}{83.9} &
  83.4 &
  \multicolumn{1}{c|}{66.4} &
  65.9 &
  \multicolumn{1}{c|}{64.1} &
  42.0 \\ \hline  
\textit{LLaMA-OA T/L Desc} &
  \multicolumn{1}{c|}{65.3} &
  65.2 &
  \multicolumn{1}{c|}{73.7} &
  73.6 &
  \multicolumn{1}{c|}{88.4} &
  48.2 \\ \hline
\textit{LLaMA-OA Memory Recall } &
  \multicolumn{1}{c|}{82.6} &
  82.1 &
  \multicolumn{1}{c|}{64.2} &
  63.8 &
  \multicolumn{1}{c|}{88.1} &
  46.8 \\ \hline
\textit{LLaMA-OA Memory Recall (OA 7)} &
  \multicolumn{1}{c|}{76.4} &
  76.3 &
  \multicolumn{1}{c|}{67.8} &
  67.9 &
  \multicolumn{1}{c|}{67.9} &
  43.0 \\ \hline  
\textit{OA Few-sample} &
  \multicolumn{1}{c|}{87.7} &
  86.9 &
  \multicolumn{1}{c|}{66.5} &
  67.3 &
  \multicolumn{1}{c|}{75.4} &
  59.8 \\ \hline  
\rowcolor[HTML]{DAE8FC} 
\multicolumn{1}{|c|}{\cellcolor[HTML]{DAE8FC}} &
  \multicolumn{2}{c|}{\cellcolor[HTML]{DAE8FC}\textbf{Rumor Stance}} &
  \multicolumn{2}{c|}{\cellcolor[HTML]{DAE8FC}\textbf{Sarcasm}} &
  \multicolumn{2}{c|}{\cellcolor[HTML]{DAE8FC}\textbf{Hate Speech}} \\ \cline{2-7} 
\rowcolor[HTML]{DAE8FC} 
\multicolumn{1}{|c|}{\multirow{-2}{*}{\cellcolor[HTML]{DAE8FC}\textbf{Model}}} &
  \multicolumn{1}{c|}{\cellcolor[HTML]{DAE8FC}\textbf{Accuracy}} &
  \textbf{F1} &
  \multicolumn{1}{c|}{\cellcolor[HTML]{DAE8FC}\textbf{Accuracy}} &
  \textbf{F1} &
  \multicolumn{1}{c|}{\cellcolor[HTML]{DAE8FC}\textbf{Accuracy}} &
  \textbf{F1} \\ \hline
\rowcolor[HTML]{DAE8FC} 
\textit{Logistic Regression} &
  \multicolumn{1}{c|}{\cellcolor[HTML]{DAE8FC}68.5} &
  40.9 &
  \multicolumn{1}{c|}{\cellcolor[HTML]{DAE8FC}76.1} &
  53.5 &
  \multicolumn{1}{c|}{\cellcolor[HTML]{DAE8FC}83.2} &
  79.2 \\ \hline
\rowcolor[HTML]{DAE8FC} 
\textit{BERT-large} &
  \multicolumn{1}{c|}{\cellcolor[HTML]{DAE8FC}\textbf{73.2}} &
  \textbf{48.2} &
  \multicolumn{1}{c|}{\cellcolor[HTML]{DAE8FC}\textbf{78.9}} &
  58.4 &
  \multicolumn{1}{c|}{\cellcolor[HTML]{DAE8FC}\textbf{84.5}} &
  \textbf{81.2} \\ \hline
\textit{GPT Basic After} &
  \multicolumn{1}{c|}{53.0} &
  36.2 &
  \multicolumn{1}{c|}{74.3} &
  \cellcolor[HTML]{C2F3D3}\textbf{65.8} &
  \multicolumn{1}{c|}{72.9} &
  77.0 \\ \hline  
\textit{GPT Basic Before} &
  \multicolumn{1}{c|}{51.5} &
  33.3 &
  \multicolumn{1}{c|}{62.9} &
  \cellcolor[HTML]{C2F3D3}\textbf{59.7} &
  \multicolumn{1}{c|}{70.4} &
  69.1 \\ \hline
\textit{GPT T/L Desc} &
  \multicolumn{1}{c|}{59.2} &
  45.7 &
  \multicolumn{1}{c|}{61.3} &
  57.9 &
  \multicolumn{1}{c|}{76.9} &
  72.1 \\ \hline
\textit{GPT Memory Recall} &
  \multicolumn{1}{c|}{40.2} &
  30.9 &
  \multicolumn{1}{c|}{52.8} &
  51.7 &
  \multicolumn{1}{c|}{71.7} &
  69.6 \\ \hline
\textit{GPT Few-sample} &
  \multicolumn{1}{c|}{40.8} &
  30.6 &
  \multicolumn{1}{c|}{68.9} &
  64.9 &
  \multicolumn{1}{c|}{74.8} &
  71.8 \\ \hline    
\textit{LLaMA-OA Basic After} &
  \multicolumn{1}{c|}{61.7} &
  29.3 &
  \multicolumn{1}{c|}{41.6} &
  41.6 &
  \multicolumn{1}{c|}{56.0} &
  55.9 \\ \hline  
\textit{LLaMA-OA Basic Before} &
  \multicolumn{1}{c|}{46.1} &
  27.9 &
  \multicolumn{1}{c|}{64.4} &
  54.8 &
  \multicolumn{1}{c|}{69.8} &
  68.2 \\ \hline
\textit{LLaMA-OA Basic (OAT 7)} &
  \multicolumn{1}{c|}{63.1} &
  35.4 &
  \multicolumn{1}{c|}{61.4} &
  38.8 &
  \multicolumn{1}{c|}{58.1} &
  58.1 \\ \hline    
\textit{LLaMA-OA T/L Desc} &
  \multicolumn{1}{c|}{56.2} &
  29.0 &
  \multicolumn{1}{c|}{75.9} &
  49.9 &
  \multicolumn{1}{c|}{75.5} &
  73.3 \\ \hline
\textit{LLaMA-OA Memory Recall} &
  \multicolumn{1}{c|}{52.4} &
  34.6 &
  \multicolumn{1}{c|}{78.1} &
  43.9 &
  \multicolumn{1}{c|}{55.4} &
  55.4 \\ \hline
\textit{LLaMA-OA Memory Recall (OA 7)} &
  \multicolumn{1}{c|}{48.8} &
  33.1 &
  \multicolumn{1}{c|}{71.9} &
  42.9 &
  \multicolumn{1}{c|}{58.7} &
  58.7 \\ \hline  
\textit{LLaMA-OA Few-sample} &
  \multicolumn{1}{c|}{28.3} &
  20.7 &
  \multicolumn{1}{c|}{71.3} &
  42.6 &
  \multicolumn{1}{c|}{70.0} &
  68.4 \\ \hline    
\end{tabular}%
}
\caption{LLMs zero-shot classification results across all prompt settings. All datasets are evaluated with accuracy and macro-F1 scores.
Green highlighted cells denote prompt settings where zero-shot LLMs beat the supervised baseline (i.e., Bert-large model fine-tuned on the training set). \textbf{Bold text} denotes the best result per task. OA 7 denotes the `OpenAssistant/oasst-sft-7-llama-30b-xor' model.}
\label{tab:results}
\end{table*}

\begin{table*}[!t]
\small
\centering
\resizebox{\textwidth}{!}{%
\begin{tabular}{|l|cccc|}
\hline
\rowcolor[HTML]{DAE8FC} 
\multicolumn{1}{|c|}{\cellcolor[HTML]{DAE8FC}} &
  \multicolumn{2}{c|}{\cellcolor[HTML]{DAE8FC}\textbf{GPT}} &
  \multicolumn{2}{c|}{\cellcolor[HTML]{DAE8FC}\textbf{LLaMA-OA}} \\ \cline{2-5} 
\rowcolor[HTML]{DAE8FC} 
\multicolumn{1}{|c|}{\multirow{-2}{*}{\cellcolor[HTML]{DAE8FC}\textbf{Synonyms}}} &
  \multicolumn{1}{c|}{\cellcolor[HTML]{DAE8FC}\textbf{Accuracy}} &
  \multicolumn{1}{c|}{\cellcolor[HTML]{DAE8FC}\textbf{F1-macro}} &
  \multicolumn{1}{c|}{\cellcolor[HTML]{DAE8FC}\textbf{Accuracy}} &
  \textbf{F1-macro} \\ \hline
\textbf{Task 1} &
  \multicolumn{4}{l|}{} \\ \hline
\rowcolor[HTML]{C2F3D3} 
Complaint / not Complaint &
  \multicolumn{1}{c|}{\cellcolor[HTML]{C2F3D3}\textbf{87.8}} &
  \multicolumn{1}{c|}{\cellcolor[HTML]{C2F3D3}\textbf{86.4}} &
  \multicolumn{1}{c|}{\cellcolor[HTML]{C2F3D3}80.1} &
  79.9 \\ \hline
Grievance / not Grievance &
  \multicolumn{1}{c|}{87.3} &
  \multicolumn{1}{c|}{85.7} &
  \multicolumn{1}{c|}{\cellcolor[HTML]{DAE8FC}82.3} &
  \cellcolor[HTML]{DAE8FC}81.9 \\ \hline
Criticism / not Criticism &
  \multicolumn{1}{c|}{80.4} &
  \multicolumn{1}{c|}{77.9} &
  \multicolumn{1}{c|}{76.7} &
  76.4 \\ \hline
Dissatisfaction / no Dissatisfaction &
  \multicolumn{1}{c|}{84.6} &
  \multicolumn{1}{c|}{83.9} &
  \multicolumn{1}{c|}{66.7} &
  66.7 \\ \hline
Discontent / no Discontent &
  \multicolumn{1}{c|}{80.7} &
  \multicolumn{1}{c|}{80.0} &
  \multicolumn{1}{c|}{55.2} &
  54.2 \\ \hline
Ensemble Majority &
  \multicolumn{1}{c|}{84.8} &
  \multicolumn{1}{c|}{83.5} &
  \multicolumn{1}{c|}{76.1} &
  76.0 \\ \hline
Ensemble All Agreed &
  \multicolumn{1}{c|}{86.8} &
  \multicolumn{1}{c|}{85.1} &
  \multicolumn{1}{c|}{\cellcolor[HTML]{DAE8FC}\textbf{84.5}} &
  \cellcolor[HTML]{DAE8FC}\textbf{83.8} \\ \hline
\textbf{Task 2} &
  \multicolumn{4}{l|}{} \\ \hline
\rowcolor[HTML]{C2F3D3} 
Pro Vaccine / Anti Vaccine / Neutral &
  \multicolumn{1}{c|}{\cellcolor[HTML]{C2F3D3}72.4} &
  \multicolumn{1}{c|}{\cellcolor[HTML]{C2F3D3}73.6} &
  \multicolumn{1}{c|}{\cellcolor[HTML]{C2F3D3}64.2} &
  63.7 \\ \hline
In Favour of the Vaccine / Against the Vaccine / Neutral &
  \multicolumn{1}{c|}{\cellcolor[HTML]{DAE8FC}73.5} &
  \multicolumn{1}{c|}{\cellcolor[HTML]{DAE8FC}74.2} &
  \multicolumn{1}{c|}{\cellcolor[HTML]{DAE8FC}\textbf{64.4}} &
  \cellcolor[HTML]{DAE8FC}\textbf{63.9} \\ \hline
Positive Sentiment / Negative Sentiment / Neutral &
  \multicolumn{1}{c|}{70.8} &
  \multicolumn{1}{c|}{70.8} &
  \multicolumn{1}{c|}{58.9} &
  52.5 \\ \hline
Belief in vaccine / not Belief in Vaccine / Neutral &
  \multicolumn{1}{c|}{\cellcolor[HTML]{DAE8FC}74.4} &
  \multicolumn{1}{c|}{\cellcolor[HTML]{DAE8FC}75.2} &
  \multicolumn{1}{c|}{61.9} &
  59.5 \\ \hline
Positive Attitude to Vaccine / Negative Attitude / Neutral &
  \multicolumn{1}{c|}{72.3} &
  \multicolumn{1}{c|}{72.3} &
  \multicolumn{1}{c|}{63.7} &
  61.3 \\ \hline
Ensemble Majority &
  \multicolumn{1}{c|}{\cellcolor[HTML]{DAE8FC}\textbf{74.7}} &
  \multicolumn{1}{c|}{\cellcolor[HTML]{DAE8FC}\textbf{75.4}} &
  \multicolumn{1}{c|}{64.2} &
  63.5 \\ \hline
\textbf{Task 3} &
  \multicolumn{1}{l|}{} &
  \multicolumn{1}{l|}{} &
  \multicolumn{1}{l|}{} &
  \multicolumn{1}{l|}{} \\ \hline
\rowcolor[HTML]{C2F3D3} 
Bragging / not Bragging &
  \multicolumn{1}{c|}{\cellcolor[HTML]{C2F3D3}84.3} &
  \multicolumn{1}{c|}{\cellcolor[HTML]{C2F3D3}\textbf{66.2}} &
  \multicolumn{1}{c|}{\cellcolor[HTML]{C2F3D3}82.8} &
  62.6 \\ \hline
Boasting / not Boasting &
  \multicolumn{1}{c|}{82.7} &
  \multicolumn{1}{c|}{65.2} &
  \multicolumn{1}{c|}{78.4} &
  60.9 \\ \hline
Showing off / not Showing off &
  \multicolumn{1}{c|}{78.8} &
  \multicolumn{1}{c|}{62.9} &
  \multicolumn{1}{c|}{\cellcolor[HTML]{DAE8FC}\textbf{88.4}} &
  56.3 \\ \hline
Self-aggrandizing / not Self-aggrandizing &
  \multicolumn{1}{c|}{81.1} &
  \multicolumn{1}{c|}{62.0} &
  \multicolumn{1}{c|}{\cellcolor[HTML]{DAE8FC}88.1} &
  60.1 \\ \hline
Excessively Proud / not Excessively Proud &
  \multicolumn{1}{c|}{75.2} &
  \multicolumn{1}{c|}{58.0} &
  \multicolumn{1}{c|}{77.9} &
  58.1 \\ \hline
Ensemble Majority &
  \multicolumn{1}{c|}{83.4} &
  \multicolumn{1}{c|}{65.4} &
  \multicolumn{1}{c|}{\cellcolor[HTML]{DAE8FC}86.0} &
  \cellcolor[HTML]{DAE8FC}\textbf{63.9} \\ \hline
Ensemble All Agreed &
  \multicolumn{1}{c|}{\cellcolor[HTML]{DAE8FC}\textbf{84.9}} &
  \multicolumn{1}{c|}{64.4} &
  \multicolumn{1}{c|}{\cellcolor[HTML]{DAE8FC}88.1} &
  59.8 \\ \hline
\textbf{Task 4} &
  \multicolumn{1}{l|}{} &
  \multicolumn{1}{l|}{} &
  \multicolumn{1}{l|}{} &
  \multicolumn{1}{l|}{} \\ \hline
\rowcolor[HTML]{C2F3D3} 
Support / Deny / Query / Comment &
  \multicolumn{1}{c|}{\cellcolor[HTML]{C2F3D3}\textbf{51.5}} &
  \multicolumn{1}{c|}{\cellcolor[HTML]{C2F3D3}\textbf{33.3}} &
  \multicolumn{1}{c|}{\cellcolor[HTML]{C2F3D3}46.1} &
  27.9 \\ \hline
Backing / Dismiss / Questioning / Comment &
  \multicolumn{1}{c|}{40.4} &
  \multicolumn{1}{c|}{30.2} &
  \multicolumn{1}{c|}{\cellcolor[HTML]{DAE8FC}52.1} &
  \cellcolor[HTML]{DAE8FC}\textbf{43.8} \\ \hline
Support / Dismiss / Questioning / Comment &
  \multicolumn{1}{c|}{39.7} &
  \multicolumn{1}{c|}{30.4} &
  \multicolumn{1}{c|}{\cellcolor[HTML]{DAE8FC}55.4} &
  \cellcolor[HTML]{DAE8FC}39.3 \\ \hline
Ensemble &
  \multicolumn{1}{c|}{41.7} &
  \multicolumn{1}{c|}{30.6} &
  \multicolumn{1}{c|}{\cellcolor[HTML]{DAE8FC}\textbf{55.5}} &
  \cellcolor[HTML]{DAE8FC}39.4 \\ \hline
\textbf{Task 5} &
  \multicolumn{4}{l|}{} \\ \hline
\rowcolor[HTML]{C2F3D3} 
Sarcasm / not Sarcasm &
  \multicolumn{1}{c|}{\cellcolor[HTML]{C2F3D3}62.9} &
  \multicolumn{1}{c|}{\cellcolor[HTML]{C2F3D3}59.7} &
  \multicolumn{1}{c|}{\cellcolor[HTML]{C2F3D3}64.4} &
  \textbf{54.8} \\ \hline
Ironic / not Ironic &
  \multicolumn{1}{c|}{\cellcolor[HTML]{DAE8FC}74.9} &
  \multicolumn{1}{c|}{\cellcolor[HTML]{DAE8FC}67.2} &
  \multicolumn{1}{c|}{63.9} &
  54.7 \\ \hline
Insincere / Sincere &
  \multicolumn{1}{c|}{\cellcolor[HTML]{DAE8FC}73.8} &
  \multicolumn{1}{c|}{\cellcolor[HTML]{DAE8FC}64.8} &
  \multicolumn{1}{c|}{\cellcolor[HTML]{DAE8FC}68.2} &
  42.7 \\ \hline
Disingenous / Genuine &
  \multicolumn{1}{c|}{\cellcolor[HTML]{DAE8FC}77.8} &
  \multicolumn{1}{c|}{\cellcolor[HTML]{DAE8FC}61.9} &
  \multicolumn{1}{c|}{56.8} &
  49.3 \\ \hline
Satire / not Satire &
  \multicolumn{1}{c|}{\cellcolor[HTML]{DAE8FC}76.9} &
  \multicolumn{1}{c|}{\cellcolor[HTML]{DAE8FC}62.8} &
  \multicolumn{1}{c|}{\cellcolor[HTML]{DAE8FC}75.2} &
  53.1 \\ \hline
Ensemble Majority &
  \multicolumn{1}{c|}{\cellcolor[HTML]{DAE8FC}74.9} &
  \multicolumn{1}{c|}{\cellcolor[HTML]{DAE8FC}\textbf{65.7}} &
  \multicolumn{1}{c|}{\cellcolor[HTML]{DAE8FC}70.5} &
  53.9 \\ \hline
Ensemble All Agreed &
  \multicolumn{1}{c|}{\cellcolor[HTML]{DAE8FC}\textbf{80.1}} &
  \multicolumn{1}{c|}{58.9} &
  \multicolumn{1}{c|}{\cellcolor[HTML]{DAE8FC}\textbf{76.9}} &
  51.2 \\ \hline
\textbf{Task 6} &
  \multicolumn{4}{l|}{} \\ \hline
\rowcolor[HTML]{C2F3D3} 
Offensive / Non-offensive &
  \multicolumn{1}{c|}{\cellcolor[HTML]{C2F3D3}70.4} &
  \multicolumn{1}{c|}{\cellcolor[HTML]{C2F3D3}69.1} &
  \multicolumn{1}{c|}{\cellcolor[HTML]{C2F3D3}69.8} &
  68.2 \\ \hline
Toxic / not Toxic &
  \multicolumn{1}{c|}{64.1} &
  \multicolumn{1}{c|}{63.5} &
  \multicolumn{1}{c|}{\cellcolor[HTML]{DAE8FC}70.7} &
  67.8 \\ \hline
Abusive / not Abusive &
  \multicolumn{1}{c|}{\cellcolor[HTML]{DAE8FC}72.2} &
  \multicolumn{1}{c|}{\cellcolor[HTML]{DAE8FC}69.3} &
  \multicolumn{1}{c|}{64.8} &
  64.2 \\ \hline
Hateful / not Hateful &
  \multicolumn{1}{c|}{\cellcolor[HTML]{DAE8FC}73.9} &
  \multicolumn{1}{c|}{\cellcolor[HTML]{DAE8FC}71.2} &
  \multicolumn{1}{c|}{\cellcolor[HTML]{DAE8FC}\textbf{75.6}} &
  \cellcolor[HTML]{DAE8FC}\textbf{72.5} \\ \hline
Derogatory / not Derogatory &
  \multicolumn{1}{c|}{68.2} &
  \multicolumn{1}{c|}{66.8} &
  \multicolumn{1}{c|}{58.1} &
  58.1 \\ \hline
Ensemble Majority &
  \multicolumn{1}{c|}{\cellcolor[HTML]{DAE8FC}71.4} &
  \multicolumn{1}{c|}{\cellcolor[HTML]{DAE8FC}69.7} &
  \multicolumn{1}{c|}{\cellcolor[HTML]{DAE8FC}73.6} &
  \cellcolor[HTML]{DAE8FC}71.1 \\ \hline
Ensemble All Agreed &
  \multicolumn{1}{c|}{\cellcolor[HTML]{DAE8FC}\textit{\textbf{75.1}}} &
  \multicolumn{1}{c|}{\cellcolor[HTML]{DAE8FC}\textit{\textbf{71.6}}} &
  \multicolumn{1}{c|}{\cellcolor[HTML]{DAE8FC}75.0} &
  \cellcolor[HTML]{DAE8FC}70.6 \\ \hline
\end{tabular}
}
\caption{LLMs zero-shot classification results using synonyms across all tasks. Green highlights are the original class names. Light grey highlighted cells denote where synonyms prompt settings beat the original label. \textbf{Bold text} denotes the best result per model per task.}
\label{tab:similar}
\end{table*}

\subsection{Evaluation Metrics}

Performance results are reported using two evaluation metrics:  1) Accuracy which consists of a direct comparison between the model predictions and the ground truth label; and 2) F1-macro scores are reported for situations where accuracy may not provide an adequate representation of performance, particularly for certain imbalanced datasets, such as \textit{Bragging} and \textit{Rumour Stance}.

\subsection{Hyper-parameters}\label{sec:params}
During initial explorations, we observed that using a higher temperature (e.g., 0.8 for GPT and 2 for LLaMA-OA) results in inadequate classification performance, as it introduces more randomness in the model outputs. This suggests that higher temperature settings can cause the model outputs to be non-reproducible. Therefore in this study, we use a low temperature (i.e., 0.2)\footnote{\url{https://platform.openai.com/docs/api-reference/chat/create}}  for GPT to make the model more focused and deterministic.

For LLaMA-OA, we follow the `precise hyper-parameter setup'\footnote{\url{https://open-assistant.io/dashboard}} indicated in the OpenAssistant web interface, where the Temperature is 0.1, Top P is 0.95, Repetition Penalty is 1.2 and Top K is 50. 
%KB: I commented this out as it's already said in the paragraph above
%Our early exploratory studies showed that using low temperatures can stabilise the output of the model to facilitate the reproducible results.

For BERT-large, we set the learning rate as 2e-5, the batch size as 16, and the maximum sequence length as 256. We run all baseline models three times with different random seeds and report average results. We fine-tune BERT-large on an Nvidia RTX Titan GPU with 24GB memory and run LLaMA-OA on an Nvidia A100 GPU with 40GB memory. The inference rates of LLaMA-OA and GPT are approximately 1,200 and 3,000 samples per hour respectively. 

% For fair use of GPT-3.5-turbo, we limit the number of requests to 3,000 per hour, following the OpenAI API policy.

\subsection{Reproducibility of LLM Output}
As noted above, to ensure a consistent output, we utilise low temperature values of 0.2 and 0.1 for both GPT and LLaMA-OA. To evaluate the reproducibility of the models' output, we execute the basic prompt setting of the Complaint dataset five times for each language model. Our observations reveal that LLaMA-OA consistently generates identical outputs, whereas GPT achieves approximately 99\% similarity in its outputs. Note that we consistently run LLaMA-OA on our own servers with identical hardware described in Section~\ref{sec:params}.
% \newpage

\section{Results}
%\subsection{Round 1, Round 2 and Round 3}

The experimental results are shown in Table~\ref{tab:results} and Table~\ref{tab:similar}. Next we discuss them in relation to each of our three research questions.

\begin{table*}[!t]
\centering
\begin{tabular}{|l|c|c|c|}
\hline
\rowcolor[HTML]{DAE8FC} 
\multicolumn{1}{|c|}{\cellcolor[HTML]{DAE8FC}\textbf{Datasets}} & \textbf{\# of Test Set} & \textbf{Average \# of Errors} & \textbf{\# of Unanimous Errors} \\ \hline
\textit{Complaint}    & 690   & 89  & 43  \\ \hline
\textit{Vaxx Stance}  & 559   & 145 & 82  \\ \hline
\textit{Bragging}     & 1,340 & 201 & 160 \\ \hline
\textit{Rumor Stance} & 1,114 & 557 & 475 \\ \hline
\textit{Sarcasm}      & 974   & 194 & 58  \\ \hline
\textit{Hate Speech}  & 3,380 & 845 & 302 \\ \hline
\end{tabular}
\caption{We conduct further error analysis on the model outputs across all datasets. \textbf{\# of Unanimous Error} denotes cases in which the LLM unanimously agrees on an incorrect answer while using different synonyms.}
\label{error}
\end{table*}

\noindent \textbf{(RQ 1) What level of zero-shot performance can LLMs achieve on social media classification tasks? How does zero-shot LLM performance compare against smaller state-of-the-art language models fine-tuned on the specific analysis task? }

In general, LLMs (GPT and LLaMA-OA) with zero-shot settings are able to achieve better results than the simple supervised Logistic Regression model. However, the traditional smaller fine-tuned language model (BERT-large) still outperforms the two LLMs on the majority of the tasks (4 out of 6 tasks). Furthermore, we observe that GPT consistently outperforms LLaMA-OA across all prompt settings and tasks when considering only the F1-macro measure. 
However, our results show that the accuracy of LLaMA-OA is better than that of GPT on some imbalanced datasets, such as `Bragging' and `Sarcasm'. This may be due to LLaMA-OA defaulting to the neutral class (labels without any specific speech act, such as `Not Bragging' and `Not Sarcastic').

GPT achieves the best predictive performance on two speech act detection downstream tasks, namely \textit{Complaint} (89.7 accuracy and 88.7 F1-macro) and Sarcasm (62.1 F1-macro). This suggests that LLMs can be employed as strong baseline models for zero-shot classification tasks.

\textcolor{black}{With respect to prompts, when the results of T/L Desc and Memory Recall are compared against Basic Instruction, it is observed that using a more complex prompt (e.g., adding label and paper information) does not necessarily improve model performance and may even introduce additional noise, leading to a degradation in performance. This indicates that adding complexity to the prompt might lead to the LLM not fully focusing on the human instructions.}

For speech act detection tasks such as \textit{Complaint} and \textit{Bragging}, the accuracy of LLMs exceeds 85\%, indicating that LLMs can potentially be used for data annotation as a way to reduce human annotation costs. Standard data annotation tasks typically rely on at least two annotators in the first round, so one of them could be replaced by an LLM. 
According to the annotation details\footnote{\url{https://github.com/sohampoddar26/covid-vax-stance/tree/main/dataset}} of the vaccine stance task \citep{poddar2022winds}, the agreement rate between the two annotators is approximately 62\%.

\noindent \textbf{(RQ 2) What are the most effective LLM prompt strategies for social media classification tasks in a zero-shot setting?}

Table~\ref{tab:results} compares different prompt complexity, and shows that the simple prompt strategy works reasonably well. For GPT, adding task and label descriptions typically achieves better results, i.e. these prompts achieved the best results on 4 out of 6 datasets as compared to other GPT prompt strategies. On the other hand, LLaMA-OA achieves mixed results. On average, for LLaMA-OA, simple prompts outperform complex counterparts. This may happen because complex prompts add additional noise to the model. 
We also note that adding a few examples to the prompt actually damages  classification performance, for both GPT and LLaMA-OA. We hypothesise that the longer prompt is affecting the model interpretation of instructions.

\begin{table*}[!t]
\centering
\begin{tabular}{|l|cc|cc|}
\hline
\rowcolor[HTML]{DAE8FC} 
\multicolumn{1}{|c|}{\cellcolor[HTML]{DAE8FC}} &
  \multicolumn{2}{c|}{\cellcolor[HTML]{DAE8FC} Basic} &
  \multicolumn{2}{c|}{\cellcolor[HTML]{DAE8FC} T/L Desc} \\ \cline{2-5} 
\rowcolor[HTML]{DAE8FC} 
\multicolumn{1}{|c|}{\multirow{-2}{*}{\cellcolor[HTML]{DAE8FC}Task}} &
  \multicolumn{1}{c|}{\cellcolor[HTML]{DAE8FC}Tokens (Sum/Mean)} &
  \multicolumn{1}{c|}{\cellcolor[HTML]{DAE8FC}\$} &
  \multicolumn{1}{c|}{\cellcolor[HTML]{DAE8FC}Tokens (Sum/Mean)} &
  \multicolumn{1}{c|}{\cellcolor[HTML]{DAE8FC}\$} \\ \hline
Rumour   & \multicolumn{1}{c|}{35k/51}  & \textless 0.1 & \multicolumn{1}{c|}{82k/119}  & 0.2 \\ \hline
Vaccine  & \multicolumn{1}{c|}{31k/127} & \textless 0.1 & \multicolumn{1}{c|}{86k/45}   & 0.2 \\ \hline
Complaint      & \multicolumn{1}{c|}{23k/33}  & \textless 0.1 & \multicolumn{1}{c|}{62k/91}   & 0.1 \\ \hline
Bragging       & \multicolumn{1}{c|}{52k/76}  & 0.1           & \multicolumn{1}{c|}{96k/140} & 0.2 \\ \hline
HateSpeech    & \multicolumn{1}{c|}{62k/90}  & 0.1           & \multicolumn{1}{c|}{94k/137}    & 0.2 \\ \hline
Sarcasm        & \multicolumn{1}{c|}{28k/41}  & \textless 0.1 & \multicolumn{1}{c|}{50k/86}   & 0.1 \\ \hline
\end{tabular}
\caption{The cost of running GPT-3.5 for each task.}
\label{tab:cost}
\end{table*}

%\subsection{Synonyms}
Table \ref{tab:similar} shows all zero-shot results when synonyms are used in prompts for all six datasets. We observe that revising prompts with synonyms can substantially improve the zero-shot performance of LLaMA-OA, except for the \textit{Bragging} dataset. 
It is worth noting that the Sarcasm dataset is the only one where the prompt using the original categories performs worse.  
This suggests that replacing original labels with synonyms allows the LLaMA-OA model to better understand the task requirements.
%This may be due to t
\textcolor{black}{The variation in the training example distribution for both GPT and LLaMA-OA could account for the observed behaviours of the models.} For example, the LLaMA-OA model might be fine-tuned on a dataset like: `\textit{[Text including offensive language] + [Category: Abusive]}'.
Therefore, we believe that it is important to test similar words in place of the original labels when designing instructions as well as use ensemble methods.

\noindent \textbf{(RQ 3) Was the pre-training corpus of the large model already inclusive of these datasets prior to the experiment (i.e., data leakage issues)?}
\textcolor{black}{To answer this question, we test different prompting strategies (e.g., by asking about the authors and task details of each paper) to explore whether the LLMs have been exposed to the dataset beforehand. In Table \ref{tab:arxiv}, we present two examples of our testing approach by directly incorporating the titles of the RumourEval \citep{derczynski-etal-2017-semeval} and Sarcasm \citep{abu-farha-etal-2022-semeval} datasets into the prompts.
Considering that LLMs are capable of recalling task details when provided with the title of an arXiv paper (i.e., memory recall), we speculate that these LLMs might be trained on these source papers, incorporating some examples alongside their corresponding labels. However, due to the opaque nature of the training corpus utilised for these LLMs, it is uncertain to what extent these datasets were included in the training data.}

\section{Error Analysis}
To better understand the limitations of LLMs, we conduct an error analysis focusing on shared errors across all synonym settings following \citep{ziems2023can}. We manually check these wrong predictions and observe that some unanimous errors \citep{ziems2023can} (i.e., when the model agreed on an incorrect answer using different synonyms) are caused by incorrect or controversial ground truth labels. We summarise the number of wrong predictions from the synonyms experiments on GPT in Table \ref{error}.

On the other hand, we observe that LLaMA-OA often defaults to the majority category, such as `not a bragging' and `not sarcasm', which leads to higher accuracy but a lower macro-F1 measure. However, considering the high accuracy of LLM zero-shot classification performance, LLMs can still be utilised as data annotation tools (combined with human efforts) for NLP downstream tasks in CSS. We can utilise LLMs for data annotation and also to identify incorrect annotations.

\section{Conclusion}
This paper explored a number of prompting strategies for the application of Large Language Models (LLMs) in computational social science tasks. It presented a range of controlled experiments that establish the efficacy of different prompt strategies on six publicly available datasets. Our main findings are summarised as follows:

\begin{itemize}
    \item Task-specific fine-tuned models generally tend to outperform LLMs in zero-shot settings. 
    % Given the compitiable results of LLMs. they can still be 
    \item More detailed and complex prompts (e.g, by adding arXiv paper title and few-samples) do not necessarily enhance classification performance.
    \item The selection of specific words or phrases as the class label can considerably affect classification outcomes.
\end{itemize}

We therefore argue that developing prompts for zero-shot classification presents a significant challenge and recommend testing different prompt configurations before proceeding with experiments, while keeping in mind the time constraints\footnote{\url{https://platform.openai.com/docs/guides/rate-limits/overview}} and financial costs associated with LLMs (see Table \ref{tab:cost}).
\section*{Limitations and Future Work}
In this paper, we assess the zero-shot text classification performance of two popular LLMs (i.e., GPT and LLaMA-OA) on six downstream tasks in CSS. \textcolor{black}{We acknowledge that further experiments on other fine-grained CSS tasks (such as Question\&Answering and Named Entity Recognition) would be beneficial in future work.} Currently, several state-of-the-art LLMs, such as GPT-4\footnote{\url{https://openai.com/gpt-4}} and Mistral \citep{jiang2023mistral}, achieve better predictive performance than those evaluated in our work. Therefore, we plan to extend our experiments in the future to include more state-of-the-art LLMs of various sizes. \textcolor{black}{Since our focus is primarily on out-of-the-box performance, we experiment with simple alterations of the prompts. Other research, e.g. \citet{arora2022ask}, has considered prompt aggregation as well as using LLMs to auto-generate prompts. We also do not explore advanced methods such as chain-of-thought prompting, which improves LM performance by encouraging it to output its intermediate reasoning steps \cite{wei2023chainofthought, suzgun2022challenging}.}

We also tried to explore potential data leakage issues \citep{ziems2023can} by testing various prompts to verify whether our test sets have been exposed to GPT and OA. However, due to the black box nature of the training datasets of these two LLMs, we are unable to confirm the presence of data leakage.

\section*{Ethics Statement}
Our work has received ethical approval from the Ethics Committee of our university and complies with the research policies of Twitter. All datasets are obtained through the links provided in the respective research papers or by requesting them directly from the authors. Furthermore, we can confirm that the data has been fully anonymized before being fed to the LLMs for model inference.

\section*{Acknowledgements}
This research is supported by a UKRI grant EP/W011212/1 ("XAIvsDisinfo: eXplainable AI Methods for Categorisation and Analysis of COVID-19 Vaccine Disinformation and Online Debates"\footnote{\url{https://gow.epsrc.ukri.org/NGBOViewGrant.aspx?GrantRef=EP/W011212/1}}) and an EU Horizon 2020 grant (agreement no.871042) ("So-BigData++: European Integrated Infrastructure for Social Mining and BigData Analytics"\footnote{\url{http://www.sobigdata.eu}}).

\section*{References}
\bibliographystyle{lrec-coling2024-natbib}
\bibliography{lrec-coling2024-example}

\begin{thebibliography}{35}
\expandafter\ifx\csname natexlab\endcsname\relax\def\natexlab#1{#1}\fi

\bibitem[{Arora et~al.(2022)Arora, Narayan, Chen, Orr, Guha, Bhatia, Chami, Sala, and R{\'e}}]{arora2022ask}
Simran Arora, Avanika Narayan, Mayee~F Chen, Laurel~J Orr, Neel Guha, Kush Bhatia, Ines Chami, Frederic Sala, and Christopher R{\'e}. 2022.
\newblock Ask me anything: A simple strategy for prompting language models.
\newblock \emph{arXiv preprint arXiv:2210.02441}.

\bibitem[{Child et~al.(2019)Child, Gray, Radford, and Sutskever}]{child2019generating}
Rewon Child, Scott Gray, Alec Radford, and Ilya Sutskever. 2019.
\newblock Generating long sequences with sparse transformers.
\newblock \emph{arXiv preprint arXiv:1904.10509}.

\bibitem[{Cotfas et~al.(2021)Cotfas, Delcea, Roxin, Ioan{\u{a}}{\c{s}}, Gherai, and Tajariol}]{cotfas2021longest}
Liviu-Adrian Cotfas, Camelia Delcea, Ioan Roxin, Corina Ioan{\u{a}}{\c{s}}, Dana~Simona Gherai, and Federico Tajariol. 2021.
\newblock The longest month: analyzing covid-19 vaccination opinions dynamics from tweets in the month following the first vaccine announcement.
\newblock \emph{Ieee Access}, 9:33203--33223.

\bibitem[{Derczynski et~al.(2017)Derczynski, Bontcheva, Liakata, Procter, Wong Sak~Hoi, and Zubiaga}]{derczynski-etal-2017-semeval}
Leon Derczynski, Kalina Bontcheva, Maria Liakata, Rob Procter, Geraldine Wong Sak~Hoi, and Arkaitz Zubiaga. 2017.
\newblock \href {https://doi.org/10.18653/v1/S17-2006} {{S}em{E}val-2017 task 8: {R}umour{E}val: Determining rumour veracity and support for rumours}.
\newblock In \emph{Proceedings of the 11th International Workshop on Semantic Evaluation ({S}em{E}val-2017)}, pages 69--76, Vancouver, Canada. Association for Computational Linguistics.

\bibitem[{Dettmers et~al.(2021)Dettmers, Lewis, Shleifer, and Zettlemoyer}]{dettmers20228bit}
Tim Dettmers, Mike Lewis, Sam Shleifer, and Luke Zettlemoyer. 2021.
\newblock 8-bit optimizers via block-wise quantization.
\newblock \emph{arXiv preprint arXiv:2110.02861}.

\bibitem[{Devlin et~al.(2019)Devlin, Chang, Lee, and Toutanova}]{devlin2019bert}
Jacob Devlin, Ming-Wei Chang, Kenton Lee, and Kristina Toutanova. 2019.
\newblock Bert: Pre-training of deep bidirectional transformers for language understanding.
\newblock In \emph{Proceedings of the 2019 Conference of the North American Chapter of the Association for Computational Linguistics: Human Language Technologies, Volume 1 (Long and Short Papers)}, pages 4171--4186.

\bibitem[{Farha et~al.(2022)Farha, Oprea, Wilson, and Magdy}]{abu-farha-etal-2022-semeval}
Ibrahim~Abu Farha, Silviu~Vlad Oprea, Steven Wilson, and Walid Magdy. 2022.
\newblock Semeval-2022 task 6: isarcasmeval, intended sarcasm detection in english and arabic.
\newblock In \emph{Proceedings of the 16th International Workshop on Semantic Evaluation (SemEval-2022)}, pages 802--814.

\bibitem[{Ganesan et~al.(2023)Ganesan, Lal, Nilsson, and Schwartz}]{ganesan2023systematic}
Adithya~V Ganesan, Yash~Kumar Lal, August~H{\aa}kan Nilsson, and H~Andrew Schwartz. 2023.
\newblock Systematic evaluation of gpt-3 for zero-shot personality estimation.
\newblock \emph{arXiv preprint arXiv:2306.01183}.

\bibitem[{Jiang et~al.(2023)Jiang, Sablayrolles, Mensch, Bamford, Chaplot, Casas, Bressand, Lengyel, Lample, Saulnier et~al.}]{jiang2023mistral}
Albert~Q Jiang, Alexandre Sablayrolles, Arthur Mensch, Chris Bamford, Devendra~Singh Chaplot, Diego de~las Casas, Florian Bressand, Gianna Lengyel, Guillaume Lample, Lucile Saulnier, et~al. 2023.
\newblock Mistral 7b.
\newblock \emph{arXiv preprint arXiv:2310.06825}.

\bibitem[{Jin et~al.(2022)Jin, Preo{\c{t}}iuc-Pietro, Do{\u{g}}ru{\"o}z, and Aletras}]{jin2022automatic}
Mali Jin, Daniel Preo{\c{t}}iuc-Pietro, A~Do{\u{g}}ru{\"o}z, and Nikolaos Aletras. 2022.
\newblock Automatic identification and classification of bragging in social media.
\newblock In \emph{Proceedings of the 60th Annual Meeting of the Association for Computational Linguistics (Volume 1: Long Papers)}, pages 3945--3959.

\bibitem[{K{\"o}pf et~al.(2023)K{\"o}pf, Kilcher, von R{\"u}tte, Anagnostidis, Tam, Stevens, Barhoum, Duc, Stanley, Nagyfi et~al.}]{kopf2023openassistant}
Andreas K{\"o}pf, Yannic Kilcher, Dimitri von R{\"u}tte, Sotiris Anagnostidis, Zhi-Rui Tam, Keith Stevens, Abdullah Barhoum, Nguyen~Minh Duc, Oliver Stanley, Rich{\'a}rd Nagyfi, et~al. 2023.
\newblock Openassistant conversations--democratizing large language model alignment.
\newblock \emph{arXiv preprint arXiv:2304.07327}.

\bibitem[{Kuzman et~al.(2023)Kuzman, Ljube{\v{s}}i{\'c}, and Mozeti{\v{c}}}]{kuzman2023chatgpt}
Taja Kuzman, Nikola Ljube{\v{s}}i{\'c}, and Igor Mozeti{\v{c}}. 2023.
\newblock Chatgpt: Beginning of an end of manual annotation? use case of automatic genre identification.
\newblock \emph{arXiv preprint arXiv:2303.03953}.

\bibitem[{Lamichhane(2023)}]{lamichhane2023evaluation}
Bishal Lamichhane. 2023.
\newblock Evaluation of chatgpt for nlp-based mental health applications.
\newblock \emph{arXiv preprint arXiv:2303.15727}.

\bibitem[{Li et~al.(2023)Li, Chan, Zhu, Pei, Ma, Liu, and Shah}]{li2023chatgpt}
Xianzhi Li, Samuel Chan, Xiaodan Zhu, Yulong Pei, Zhiqiang Ma, Xiaomo Liu, and Sameena Shah. 2023.
\newblock Are chatgpt and gpt-4 general-purpose solvers for financial text analytics? a study on several typical tasks.
\newblock In \emph{Proceedings of the 2023 Conference on Empirical Methods in Natural Language Processing: Industry Track}, pages 408--422.

\bibitem[{Longpre et~al.(2023)Longpre, Hou, Vu, Webson, Chung, Tay, Zhou, Le, Zoph, Wei et~al.}]{longpreFlanCollectionDesigning2023}
Shayne Longpre, Le~Hou, Tu~Vu, Albert Webson, Hyung~Won Chung, Yi~Tay, Denny Zhou, Quoc~V Le, Barret Zoph, Jason Wei, et~al. 2023.
\newblock The flan collection: Designing data and methods for effective instruction tuning.
\newblock \emph{arXiv preprint arXiv:2301.13688}.

\bibitem[{Mitchell et~al.(2023)Mitchell, Lee, Khazatsky, Manning, and Finn}]{mitchell2023detectgpt}
Eric Mitchell, Yoonho Lee, Alexander Khazatsky, Christopher~D Manning, and Chelsea Finn. 2023.
\newblock Detectgpt: Zero-shot machine-generated text detection using probability curvature.
\newblock \emph{arXiv preprint arXiv:2301.11305}.

\bibitem[{Mu et~al.(2023)Mu, Jin, Grimshaw, Scarton, Bontcheva, and Song}]{mu2023vaxxhesitancy}
Yida Mu, Mali Jin, Charlie Grimshaw, Carolina Scarton, Kalina Bontcheva, and Xingyi Song. 2023.
\newblock Vaxxhesitancy: A dataset for studying hesitancy towards covid-19 vaccination on twitter.
\newblock In \emph{Proceedings of the International AAAI Conference on Web and Social Media}, volume~17, pages 1052--1062.

\bibitem[{Olshtain and Weinbach(1987)}]{olshtain198710}
Elite Olshtain and Liora Weinbach. 1987.
\newblock 10. complaints: A study of speech act behavior among native and non-native speakers of hebrew.
\newblock In \emph{The pragmatic perspective}, page 195. John Benjamins.

\bibitem[{Ouyang et~al.(2022)Ouyang, Wu, Jiang, Almeida, Wainwright, Mishkin, Zhang, Agarwal, Slama, Ray et~al.}]{ouyangTrainingLanguageModels2022}
Long Ouyang, Jeffrey Wu, Xu~Jiang, Diogo Almeida, Carroll Wainwright, Pamela Mishkin, Chong Zhang, Sandhini Agarwal, Katarina Slama, Alex Ray, et~al. 2022.
\newblock Training language models to follow instructions with human feedback.
\newblock \emph{Advances in Neural Information Processing Systems}, 35:27730--27744.

\bibitem[{Poddar et~al.(2022)Poddar, Mondal, Misra, Ganguly, and Ghosh}]{poddar2022winds}
Soham Poddar, Mainack Mondal, Janardan Misra, Niloy Ganguly, and Saptarshi Ghosh. 2022.
\newblock Winds of change: Impact of covid-19 on vaccine-related opinions of twitter users.
\newblock In \emph{Proceedings of the International AAAI Conference on Web and Social Media}, volume~16, pages 782--793.

\bibitem[{Preo{\c{t}}iuc-Pietro et~al.(2019)Preo{\c{t}}iuc-Pietro, Gaman, and Aletras}]{preoctiuc2019automatically}
Daniel Preo{\c{t}}iuc-Pietro, Mihaela Gaman, and Nikolaos Aletras. 2019.
\newblock Automatically identifying complaints in social media.
\newblock In \emph{Proceedings of the 57th Annual Meeting of the Association for Computational Linguistics}, pages 5008--5019.

\bibitem[{Reiss(2023)}]{reiss2023testing}
Michael~V Reiss. 2023.
\newblock Testing the reliability of chatgpt for text annotation and classification: A cautionary remark.
\newblock \emph{arXiv preprint arXiv:2304.11085}.

\bibitem[{Sanh et~al.(2021)Sanh, Webson, Raffel, Bach, Sutawika, Alyafeai, Chaffin, Stiegler, Scao, Raja et~al.}]{sanhMultitaskPromptedTraining2022}
Victor Sanh, Albert Webson, Colin Raffel, Stephen~H Bach, Lintang Sutawika, Zaid Alyafeai, Antoine Chaffin, Arnaud Stiegler, Teven~Le Scao, Arun Raja, et~al. 2021.
\newblock Multitask prompted training enables zero-shot task generalization.
\newblock \emph{arXiv preprint arXiv:2110.08207}.

\bibitem[{Su et~al.(2023)Su, Zhuo, Wang, and Nakov}]{su2023detectllm}
Jinyan Su, Terry~Yue Zhuo, Di~Wang, and Preslav Nakov. 2023.
\newblock Detectllm: Leveraging log rank information for zero-shot detection of machine-generated text.
\newblock \emph{arXiv preprint arXiv:2306.05540}.

\bibitem[{Suzgun et~al.(2022)Suzgun, Scales, Sch{\"a}rli, Gehrmann, Tay, Chung, Chowdhery, Le, Chi, Zhou et~al.}]{suzgun2022challenging}
Mirac Suzgun, Nathan Scales, Nathanael Sch{\"a}rli, Sebastian Gehrmann, Yi~Tay, Hyung~Won Chung, Aakanksha Chowdhery, Quoc~V Le, Ed~H Chi, Denny Zhou, et~al. 2022.
\newblock Challenging big-bench tasks and whether chain-of-thought can solve them.
\newblock \emph{arXiv preprint arXiv:2210.09261}.

\bibitem[{T{\"o}rnberg(2023)}]{törnberg2023chatgpt4}
Petter T{\"o}rnberg. 2023.
\newblock Chatgpt-4 outperforms experts and crowd workers in annotating political twitter messages with zero-shot learning.
\newblock \emph{arXiv preprint arXiv:2304.06588}.

\bibitem[{Touvron et~al.(2023)Touvron, Lavril, Izacard, Martinet, Lachaux, Lacroix, Rozi{\`e}re, Goyal, Hambro, Azhar et~al.}]{touvron2023llama}
Hugo Touvron, Thibaut Lavril, Gautier Izacard, Xavier Martinet, Marie-Anne Lachaux, Timoth{\'e}e Lacroix, Baptiste Rozi{\`e}re, Naman Goyal, Eric Hambro, Faisal Azhar, et~al. 2023.
\newblock Llama: Open and efficient foundation language models.
\newblock \emph{arXiv preprint arXiv:2302.13971}.

\bibitem[{Wang and Chang(2022)}]{wang2022toxicity}
Yau-Shian Wang and Yingshan Chang. 2022.
\newblock Toxicity detection with generative prompt-based inference.
\newblock \emph{arXiv preprint arXiv:2205.12390}.

\bibitem[{Waseem et~al.(2017)Waseem, Davidson, Warmsley, and Weber}]{waseem2017understanding}
Zeerak Waseem, Thomas Davidson, Dana Warmsley, and Ingmar Weber. 2017.
\newblock Understanding abuse: A typology of abusive language detection subtasks.
\newblock In \emph{Proceedings of the First Workshop on Abusive Language Online}, pages 78--84.

\bibitem[{Waseem and Hovy(2016)}]{waseem-hovy:2016:N16-2}
Zeerak Waseem and Dirk Hovy. 2016.
\newblock \href {http://www.aclweb.org/anthology/N16-2013} {Hateful symbols or hateful people? predictive features for hate speech detection on twitter}.
\newblock In \emph{Proceedings of the NAACL Student Research Workshop}, pages 88--93, San Diego, California. Association for Computational Linguistics.

\bibitem[{Wei et~al.(2021)Wei, Bosma, Zhao, Guu, Yu, Lester, Du, Dai, and Le}]{weiFINETUNEDLANGUAGEMODELS2022}
Jason Wei, Maarten Bosma, Vincent~Y Zhao, Kelvin Guu, Adams~Wei Yu, Brian Lester, Nan Du, Andrew~M Dai, and Quoc~V Le. 2021.
\newblock Finetuned language models are zero-shot learners.
\newblock \emph{arXiv preprint arXiv:2109.01652}.

\bibitem[{Wei et~al.(2022)Wei, Wang, Schuurmans, Bosma, brian ichter, Xia, Chi, Le, and Zhou}]{wei2023chainofthought}
Jason Wei, Xuezhi Wang, Dale Schuurmans, Maarten Bosma, brian ichter, Fei Xia, Ed~H. Chi, Quoc~V Le, and Denny Zhou. 2022.
\newblock \href {https://openreview.net/forum?id=_VjQlMeSB_J} {Chain of thought prompting elicits reasoning in large language models}.
\newblock In \emph{Advances in Neural Information Processing Systems}.

\bibitem[{Wu et~al.(2023)Wu, Tucker, Nagler, and Messing}]{wu2023large}
Patrick~Y Wu, Joshua~A Tucker, Jonathan Nagler, and Solomon Messing. 2023.
\newblock Large language models can be used to estimate the ideologies of politicians in a zero-shot learning setting.
\newblock \emph{arXiv preprint arXiv:2303.12057}.

\bibitem[{Ziems et~al.(2023)Ziems, Held, Shaikh, Chen, Zhang, and Yang}]{ziems2023can}
Caleb Ziems, William Held, Omar Shaikh, Jiaao Chen, Zhehao Zhang, and Diyi Yang. 2023.
\newblock Can large language models transform computational social science?
\newblock \emph{arXiv preprint arXiv:2305.03514}.

\bibitem[{Zubiaga et~al.(2018)Zubiaga, Aker, Bontcheva, Liakata, and Procter}]{zubiaga2018detection}
Arkaitz Zubiaga, Ahmet Aker, Kalina Bontcheva, Maria Liakata, and Rob Procter. 2018.
\newblock Detection and resolution of rumours in social media: A survey.
\newblock \emph{ACM Computing Surveys (CSUR)}, 51(2):1--36.

\end{thebibliography}

\appendix
% \newpage
\section*{Appendix}

\begin{table*}[!t]
\centering
\resizebox{\textwidth}{!}{%
\begin{tabular}{|ll|}
\hline
\multicolumn{2}{|l|}{\cellcolor[HTML]{DAE8FC}\textbf{Task \& Prompt}} \\ \hline
\multicolumn{1}{|l|}{Rumour Stance} &
  Do you know this paper: SemEval-2017 Task 8: RumourEval: Determining rumour veracity and support for rumours? \\ \hline
\multicolumn{1}{|l|}{Sarcasm} &
  Do you know this paper: Semeval-2022 task 6: isarcasmeval, intended sarcasm detection in English and Arabic? \\ \hline
\multicolumn{2}{|l|}{\cellcolor[HTML]{DAE8FC}\textbf{Output}} \\ \hline
\multicolumn{1}{|l|}{GPT (Rumour)} &
  \begin{tabular}[c]{@{}l@{}}Yes, I'm familiar with the paper titled SemEval-2017 Task 8: RumourEval: Determining rumour veracity and support \\ for rumours. This paper is related to the SemEval-2017 workshop, specifically Task 8, which focused on rumour detection \\ and veracity classification on social media platforms. Participants were provided with a dataset containing rumours along \\ with associated tweets that either supported or denied the rumours. The task involved classifying the veracity of the rumours \\ as true, false, or unverified, and also identifying the tweets that provided support or  denial for the rumours.\end{tabular} \\ \hline
\multicolumn{1}{|l|}{LLaMA-OA (Rumour)} &
  \begin{tabular}[c]{@{}l@{}}Yes, I am aware of the paper titled SemEval-2017 Task 8: RumourEval: Determining rumour veracity and support for \\ rumours. This paper presents the results of the RumourEval task held as part of the 2017 edition of the SEMEVAL \\ workshop. The task involved developing algorithms for determining the veracity of rumors and identifying sources that \\ supported or refuted them using Twitter data.\end{tabular} \\ \hline
\multicolumn{1}{|l|}{GPT (Sarcasm)} &
  \cellcolor[HTML]{B5F5F2}\begin{tabular}[c]{@{}l@{}}I apologise for the confusion, but as an AI language model, my training only goes up until September 2021, and I do not \\ have access to specific details about SemEval-2022 or any subsequent tasks. The last SemEval workshop I'm aware of is \\ SemEval-2021.\end{tabular} \\ \hline
\multicolumn{1}{|l|}{LLaMA-OA (Sarcasm)} &
  \begin{tabular}[c]{@{}l@{}}Semeval-2022 Task 6: ISARCASMEVAL is a subtask focusing specifically on the identification of ironic and sarcastic \\ expressions in English and Arabic text. Irony and sarcasm are common forms of expression found in both written and \\ spoken communication, but they pose significant challenges for automatic sentiment analysis due to their subtle nature \\ and dependence on contextual cues.\end{tabular} \\ \hline
\end{tabular}
}
\caption{Ask LLMs if they can recall the original paper.}
\label{tab:arxiv}
\end{table*}

\begin{table*}[!t]
\resizebox{\textwidth}{!}{
% \small
\begin{tabular}{|l|l|}
\hline
\rowcolor[HTML]{ECF4FF} 
\textbf{Tasks} &
  \multicolumn{1}{c|}{\cellcolor[HTML]{ECF4FF}\textbf{Prompt with Task Description}} \\ \hline
Complaint &
  \begin{tabular}[c]{@{}l@{}}Complaining is a basic speech act used to express a negative mismatch between reality and expectations \\ towards a state of affairs, product, organization or event. Key to the definition of complaints is the expression \\ of the breach of expectations.\end{tabular} \\ \hline
Vaccine Stance &
  \begin{tabular}[c]{@{}l@{}}Pro-vaccine tweets express a positive opinion regarding vaccination. Anti-vaccine tweets express a negative \\ opinion towards COVID-19 vaccination. Neutral includes news related to vaccine development, questions \\ about the vaccine, or informative tweets concerning vaccination without a clear opinion.\end{tabular} \\ \hline
Rumour Stance &
  \begin{tabular}[c]{@{}l@{}}Support: the author of the response supports the veracity of the rumour. Deny: the author of the response \\ denies the veracity of the rumour. Query: the author of the response asks for additional evidence in relation to \\ the veracity of the rumour. Comment: the author of the response makes their own comment without a clear \\ contribution to assessing the veracity of the rumour.\end{tabular} \\ \hline
Hate Speech &
  \begin{tabular}[c]{@{}l@{}}A tweet is offensive if it: 1. uses a sexist or racial slur. 2. attacks a minority. 3. seeks to silence a minority. \\ 4. criticizes a minority (without a well founded argument). 5. promotes, but does not directly use, hate \\ speech or violent crime. 6. criticizes a minority and uses a straw man argument. 7. blatantly misrepresents\\ truth or seeks to distort views on a minority with unfounded claims. 8. shows support of problematic hash \\ tags. E.g. “\#BanIslam”, “\#whoriental”, “\#whitegenocide”. 9. negatively stereotypes a minority. 10. defends \\ xenophobia or sexism. 11. contains a screen name that is offensive, as per the previous criteria, the tweet is \\ ambiguous (at best), and the tweet is on a topic that satisfies any of the above criteria.\end{tabular} \\ \hline
Sarcasm &
  \begin{tabular}[c]{@{}l@{}}Sarcasm is a form of verbal irony that occurs when there is a discrepancy between the literal and intended \\ meanings of an utterance. Through this discrepancy, the speaker expresses their position towards a prior \\ proposition, often in the form of surface contempt or derogation.\end{tabular} \\ \hline
Bragging &
  \begin{tabular}[c]{@{}l@{}}Bragging is a speech act which explicitly or implicitly attributes credit to the speaker for some ‘good’ \\ (possession, accomplishment, skill, etc.) which is positively valued by the speaker and the potential audience.\\ As such, bragging includes announcements of accomplishments, and explicit positive evaluations of some \\ aspect of self. A bragging statement should clearly express what the author is bragging about (i.e. the target \\ of bragging).\end{tabular} \\ \hline
\end{tabular}
}
\caption{Task descriptions used for the prompting strategy \textbf{`Task and Label Description (T/L Desc)'}.}
\label{tab:taskdescription}
\end{table*}

\begin{table*}[]
\resizebox{\textwidth}{!}{%
\begin{tabular}{|l|l|}
\hline
\rowcolor[HTML]{ECF4FF} 
\textbf{Tasks} &
  \multicolumn{1}{c|}{\cellcolor[HTML]{ECF4FF}\textbf{Memory Recall Prompt}} \\ \hline
Complaint &
  \begin{tabular}[c]{@{}l@{}}Recall paper: (Automatically identifying complaints in social media). Identify whether a tweet is a customer \\ complaint or a non-complaint. Only reply 'Complaint' or Non-complaint'.\end{tabular} \\ \hline
Vaccine Stance &
  \begin{tabular}[c]{@{}l@{}}Recall paper: (The Longest Month: Analyzing COVID-19 Vaccination Opinions Dynamics From Tweets in \\ the Month Following the First Vaccine Announcement). Annotate a tweet into one of three stance categories: \\ pro vaccine, anti vaccine, or neutral. Only reply the stance.\end{tabular} \\ \hline
Rumour Stance &
  \begin{tabular}[c]{@{}l@{}}Recall paper: (SemEval-2017 Task 8: RumourEval: Determining rumour veracity and support for rumours). \\ Classify a tweet into one of four stance categories: support, deny, query, or comment. Only reply the stance.\end{tabular} \\ \hline
Hate Speech &
  \begin{tabular}[c]{@{}l@{}}Recall paper: (Hateful symbols or hateful people? predictive features for hate speech detection on twitter.) \\ Annotate whether a tweet is hateful or not hateful. Only reply hateful or not hateful.\end{tabular} \\ \hline
Sarcasm &
  \begin{tabular}[c]{@{}l@{}}Recall paper: (Semeval-2022 task 6: isarcasmeval, intended sarcasm detection in English and Arabic.) \\ Annotate whether a tweet is sarcastic or not sarcastic. Only reply sarcastic or not sarcastic.\end{tabular} \\ \hline
Bragging &
  \begin{tabular}[c]{@{}l@{}}Recall paper: (Automatic Identification and Classification of Bragging in Social Media) Identify whether or \\ not a tweet includes a bragging statement. Only reply yes or no.\end{tabular} \\ \hline
\end{tabular}
}
\caption{Prompts for the Memory Recall strategy.}
\label{tab:memory recall}
\end{table*}

\begin{table*}[]
\resizebox{\textwidth}{!}{
\begin{tabular}{|l|l|}
\hline
\rowcolor[HTML]{ECF4FF} 
\textbf{Tasks} &
  \multicolumn{1}{c|}{\cellcolor[HTML]{ECF4FF}\textbf{Prompt with Few-sample}} \\ \hline
Complaint &
  \begin{tabular}[c]{@{}l@{}}Complaint: @USER @USER give the timeline by which I'll receive my cashback which I should have \\ received by 15th October 2017. \\ Not Complaint: I just gave 5 stars to Nancy at @USER for the great service I received!\end{tabular} \\ \hline
Vaccine Stance &
  \begin{tabular}[c]{@{}l@{}}Pro Vaccine: This is very encouraging!! I've been hoping for a vaccine in 1Q21 so I can confidently \\ travel to my favorite nephew's wedding in northern California in April. \\ Anti Vaccine: This is why I have no faith that in Covid vaccines - Covid only harms those with \\ compromised immune response, the very same people for whom vaccines don't seem to work (because \\ their immune system isn't working properly). \\ Neutral: Medical supplies across specialties are allocated under two budget items totalling RM4.29 \\ billion in MOH’s Specific Programmes in operating expenditure of Budget 2021; Covid-19 vaccines \\ are also not listed for MOH under the separate Covid-19 Fund\end{tabular} \\ \hline
Rumour Stance &
  \begin{tabular}[c]{@{}l@{}}Support: @USER @USER @USER @USER yeah i feel really sorry for them \\ Deny: @USER I never called uber PT . Everyone is having a go at Uber but not PT ... We own it , we \\ shouldn't have to pay in desperate times \\ Query: @USER @USER Ironic since all the i witnesses say the officer was white . Now it is the black \\ officer Darren Wilson who shot ? ?\\ Comment: @USER @USER Uber is covering the cost of all rides , Uber is still paying drivers higher \\ fares to encourage them to do pickups.\end{tabular} \\ \hline
Hate Speech &
  \begin{tabular}[c]{@{}l@{}}Hateful: @USER Tell it to the 120 million Africans that Islam murdered. URL\\ Not Hateful: @USER @USER doesn't look like I am.\end{tabular} \\ \hline
Sarcasm &
  \begin{tabular}[c]{@{}l@{}}Sarcastic: I love days when Rob works short call and is only at the hospital for *checks watch* 13 hours.\\ Not Sarcastic: I got stop putting on glitter flowers I'd like to ad red.\end{tabular} \\ \hline
Bragging &
  \begin{tabular}[c]{@{}l@{}}Bragging: Come watch me and @USER face off in 2K best of 3 series \#braggingrights @USER you next \\ boiiii :flushed\_face: :hot\_face:.\\ Not Bragging: I have completed survey on NaMo App.\end{tabular} \\ \hline
\end{tabular}
}
\caption{Examples used for the prompting strategy \textbf{`Few-sample'}.}
\label{tab:f-sample}
\end{table*}

\end{document}